\documentclass{vgtc}   
\ifpdf               
  \pdfoutput=1\relax   
  \usepackage{graphicx}        \DeclareGraphicsExtensions{.pdf,.png,.jpg,.jpeg} 
\else
  \usepackage{graphicx}             
  \DeclareGraphicsExtensions{.eps}  
\fi

\graphicspath{{figures/}{pictures/}{images/}{./}}
\usepackage{microtype}       
\PassOptionsToPackage{warn}{textcomp}
\usepackage{textcomp}              
\usepackage{mathptmx}             
\usepackage{times}              
        
\usepackage{cite}                
\usepackage{tabu}             
\usepackage{hyperref}
\usepackage{booktabs}            
\usepackage{xcolor}
\usepackage{pgfplots, pgfplotstable}
\definecolor{firstcolor}{HTML}{55a868}
\definecolor{secondcolor}{HTML}{cfcfcf}
\definecolor{thirdcolor}{HTML}{c44e52}
\usepackage{enumitem}
\usepackage{tikz}

\onlineid{0}
\vgtccategory{Research}

\vgtcinsertpkg

\title{OCTOPUS: Open-vocabulary Content Tracking and Object Placement Using Semantic Understanding in Mixed Reality}

\author{Luke Yoffe\thanks{Yoffe and Sharma contributed equally to this work}*
\and Aditya Sharma*
\and Tobias Höllerer
}
\affiliation{
\scriptsize University of California, Santa Barbara}

\abstract{One key challenge in augmented reality is the placement of virtual content in natural locations. Existing automated techniques are only able to work with a closed-vocabulary, fixed set of objects. In this paper, we introduce a new open-vocabulary method for object placement. Our eight-stage pipeline leverages recent advances in segmentation models, vision-language models, and LLMs to place any virtual object in any AR camera frame or scene. In a preliminary user study, we show that our method performs at least as well as human experts 57$\%$ of the time.
} 

\CCScatlist{
  \CCScatTwelve{Semantic Content Placement}{Augmented Reality}{Vision and Language}{Large Language Models}
}

\begin{document}
\maketitle

\section{Introduction}
\begin{figure}[t]
  \centering
  \includegraphics[width=1\linewidth]{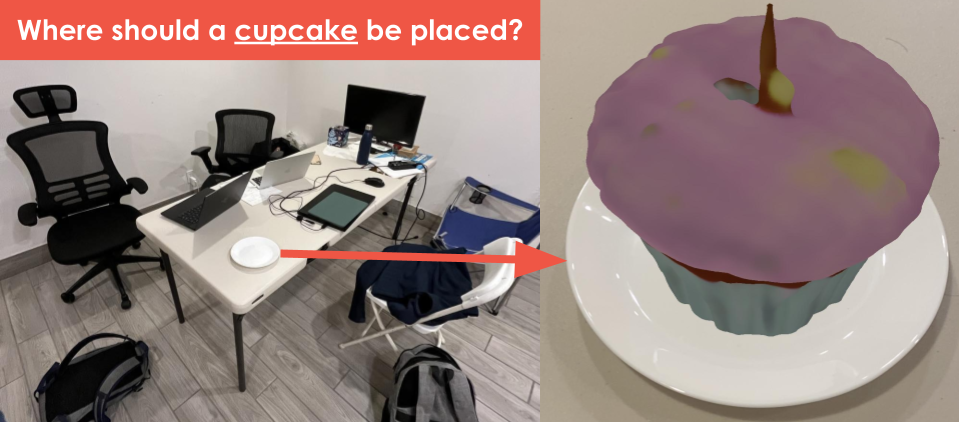}
  \caption{
  Result of our proposed method OCTOPUS, where we take an image of the scene (\emph{left}) and determine a natural location for, e.g., a \texttt{\textbf{cupcake}} to be placed, such as the plate in the scene (\emph{right}).}
  \label{fig:teaser}
\end{figure}

Augmented reality (AR) promises to seamlessly blend digital content with the real world, which requires placing virtual content in natural locations. For example, in \autoref{fig:teaser}, the plate is a natural location for a virtual cupcake to be placed. Currently, automated placement techniques do exist, but they are not able to work with arbitrary objects and scenes as the underlying machine learning models are closed-vocabulary. This means that the models are only able to handle a fixed set of words. Open-vocabulary models, on the other hand are able to adapt to words not seen during training. We combined several such models together to arrive at OCTOPUS, an eight-step method described in \autoref{method}. OCTOPUS accepts as input an image of a scene and a text description of a virtual object, and determines where in the scene the object should be placed. 

\section{Related Work}
In the context of automated virtual content placement, two interpretations of {\em natural} have been explored. First, virtual content should follow the laws of physics and be aligned with planar surfaces \cite{nuernberger2016snaptoreality}. To evaluate object placement from a physical perspective, Rafi et al. \cite{PredART} introduced a framework that predicts human ratings for object placements. {\em Natural} can also be interpreted from a semantic perspective, which is this paper's focus. Cheng et al. \cite{SemanticAdapt} and Lang et al.\cite{AgentPlacement} used scene semantics to place virtual interface elements (such as virtual screens) and virtual agents respectively. These works focused on placing specific objects, whereas our goal is to create a single pipeline that can place any object with no dedicated training.

\section{Method}\label{method}
Our virtual content placement pipeline is divided into eight steps:

\noindent {\bf  1. Input: }
As input, our pipeline expects an image, which could be a camera frame from an AR application, and a text prompt naming the object to be placed.

\noindent {\bf 2. Image Segmentation: }
Next, we detect all potential objects in the scene using Segment Anything Model (SAM) \cite{sam}. SAM outputs many image regions that could potentially contain an object. We determine the bounding box for every region and retrieve the corresponding image patches.

\noindent {\bf 3. Image Captioning: }
To identify the objects in each image patch, we leverage clip-text-decoder\cite{cliptd}, an open source project. Clip-text-decoder generates a text caption for an image by encoding the image into an embedding (using CLIP: Contrastive Language-Image Pretraining \cite{clip}) and then decoding the embedding into text. For every image patch, we run clip-text-decoder to generate a caption.

\noindent {\bf 4. Noun Extraction: }
Next, we list all objects referenced in each caption, since any of them could be sites to place the virtual object. All objects are nouns, so we use English Part-of-Speech tagging in Flair \cite{POS} to assign a part of speech to each word in each caption, keeping only the words marked as nouns.

\noindent {\bf 5. Noun Filtration: }
Some of the nouns found may have been misidentified and we now filter out such cases. We use the Vision-and-language Transformer \cite{vilt} (ViLT) model, which can perform visual question answering. We craft the question, ``\texttt{Is there a \{}{\it noun}\texttt{\} in the image?}", for each noun from Step 4. We include ``\texttt{floor}" in the set of nouns as clip-text-decoder often failed to mention it. We then feed ViLT the image and ask the question for each noun, keeping the nouns that led ViLT to output ``\texttt{yes}".

\noindent {\bf 6. Noun Selection: }
To take advantage of LLM reasoning-like capabilities\cite{sparks}, we use prompt engineering on OpenAI's GPT-4 in order to select the noun where the object should be placed. We arrived at the following prompt for GPT-4: ``\texttt{Give a one word response to fill in the blank using only one of these options: \{{\it list of nouns}\texttt{\}}. The \{}{\it object}\texttt{\} was located on the \underline{\hspace{0.75cm}}.}", where the list of nouns is provided by Step 5. GPT-4 returns the most likely choice.

\noindent {\bf 7. Location in Image: }\label{loc-image}
Next, we use CLIPSeg \cite{clipseg} to locate the selected noun in the image by feeding it the image and noun from GPT-4. CLIPSeg generates a heatmap indicating the similarity between each region in the image and the provided text prompt. We identify the brightest location $(x,y)$ in the heatmap, which is the pixel in the image most related to the input noun.

\noindent {\bf 8. Location in Scene: }\label{loc-scene}
After determining the 2D $(x,y)$ location in the image, our last step is to find the corresponding 3D $(x, y, z)$ position in the scene, which is where the virtual object will be placed in augmented reality. To accomplish this, we employ ray casting into the scene (modeled, e.g., by ARKit or ARCore). We place the object at the first intersection point between the ray and the scene.

\section{Results}
This chaining of semantic ML technologies provides remarkably robust performance with general scenes and objects. To measure the performance of the OCTOPUS method, we designed an experiment that compares four placement methods: (1) experts' {\em natural} placements, (2) experts' {\em unnatural} placements, (3) {\em random} placements, and (4) OCTOPUS placements.

\noindent{\bf Experiment Setup:} In order to perform the experiment, a representative set of objects and images to test with was required. We created an unbiased diverse list of 15 objects that are commonly found indoors \texttt{(apple, cake, cup, plate, vase, stool, painting, lamp, book, bag, computer, pencil, shoes, cushion, cat)}. We randomly sampled 100 indoor scene images from the NYU Depth Dataset\cite{nyudepth} and Sun3D\cite{sun3d} to annotate with object placement locations.

\noindent{\bf Annotation:} We had two experts annotate each of the 100 images with a {\em natural} and {\em unnatural} location to place each of the 15 objects. For example, in \autoref{fig:teaser}, it would be {\em natural} to place a cupcake on the plate, but {\em unnatural} to place a cupcake on the floor. Any objects that were deemed unsuitable or irrelevant for a specific image were excluded from further analysis throughout the experiment for that particular image (this happened in 573 of the 1,500 image-object combinations). We also generated placement coordinates using random point selection, and finally the OCTOPUS model. In the end, we arrived at 927 object location-image pairs for each of the four placement methods.

\begin{figure}[t]
\begin{tikzpicture}[scale=0.72]
\pgfplotstableread{
Label   First   Second  Third Color
160      0.980     0.000     0.02   Natural\ vs\ Unnatural
20       0.960     0.040     0.00   Natural\ vs\ Random
20       0.860     0.140     0.000  OCTOPUS\ vs\ Unnatural
10       0.860     0.120     0.02   OCTOPUS\ vs\ Random
160      0.060     0.51      0.43   OCTOPUS\ vs\ Natural

}\datatable

\begin{axis}[
    xbar stacked,
    xmin=0,
    xmax=1,
    ymin=-0.5,
    ymax=4.5,
    ytick=data,
    y=0.35cm,
    yticklabels from table={\datatable}{Color},
    legend style={at={(1.05,1)},anchor=north west},
]

\addplot [fill=firstcolor] table [x=First, y expr=\coordindex] {\datatable};  
\addplot [fill=secondcolor]table [x=Second, y expr=\coordindex] {\datatable};
\addplot [fill=thirdcolor] table [x=Third, y expr=\coordindex] {\datatable};
\legend{Win, Tie, Lose}
\end{axis}
\end{tikzpicture}
\caption{Evaluation results. ``Win" and ``Lose" indicate the proportion of time that the method listed first won or lost. ``Tie" indicates the proportion of time that both methods' placements were equally natural.
}
\label{fig:placements}
\end{figure}
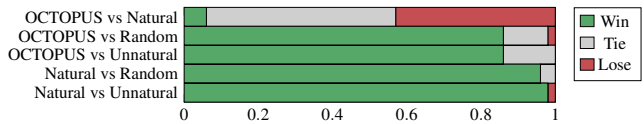

\noindent{\bf Evaluation:}\label{results_section} We compared two methods against each other at a time, omitting the comparison of unnatural and random placements, resulting in the five comparisons depicted in \autoref{fig:placements}. In each one, evaluators were told what object was to be placed and were shown two images side-by-side. Both images were annotated with a red circle indicating the proposed placement location. The evaluators then selected which placement location was superior or declared a tie if both locations were deemed equally appropriate for the object in question. The evaluators did not know which method produced each placement location. We repeated this judgment task with 100 randomly sampled object-image pairs for the OCTOPUS vs. natural comparison, and 50 for each of the remaining four method duels.

The results, shown in \autoref{fig:placements}, reveal that 57\% of the time, OCTOPUS selected a location at least as natural as the human expert selecting a {\em natural} location. The experts' natural locations won over the random and unnatural locations 96\% and 98\% of the time respectively, which confirms that they were indeed appropriate locations. OCTOPUS also won over the random and unnatural locations the vast majority of the time, demonstrating that it is tailored to human preferences and not far off from the experts' {\em natural} placements.

\section{Limitations and Future Work}
While our method generally places objects naturally, it has limitations. First, it takes around 30 seconds to generate a single placement location on an NVIDIA RTX A4000, which could be impractical in real-world applications, in particular when making live queries with AR cameras. Additionally, while our method selects the best entity for virtual object placement, it does not consider where on the entity would appear the most natural. For example, OCTOPUS could place a painting on a wall, but would follow CLIPSeg's highest heatmap response for \texttt{wall}, which may not match the natural eye level placement for paintings.

To support future work, we believe that an automated metric to determine the quality of semantic object placement would be of great value, as it could replace costly user studies.

\section{Conclusion}
We present OCTOPUS, a technique for placing virtual content in augmented reality. OCTOPUS takes as input an image and a text description of an object to be placed in the scene. It then detects entities in the image and uses LLM reasoning to determine the best entity for the object to be placed on. Lastly, it locates and places the object on the selected entity. The entire OCTOPUS pipeline is open-vocabulary, meaning it can be used to place any object in any scene out of the box, without any fine tuning. We find in preliminary evaluation that over 57$\%$ of the time OCTOPUS places objects at least as naturally as human experts.

\bibliographystyle{abbrv-doi-narrow}
\bibliography{template}
\end{document}